\documentclass[]{fairmeta}

\usepackage[utf8]{inputenc} 
\usepackage[T1]{fontenc}    
\usepackage{url}            
\usepackage{nicefrac}     
\usepackage{microtype}    

\usepackage{wrapfig}
\usepackage{graphicx}
\usepackage{amsmath}
\usepackage{amssymb}
\usepackage{pifont}
\usepackage{booktabs}
\usepackage{array}
\usepackage{tabularx}
\usepackage{graphicx, amsmath, multirow, overpic, textpos} 
\usepackage{colortbl}
\usepackage[linesnumbered,ruled,vlined]{algorithm2e}
\SetKwInOut{Parameter}{Parameters}
\usepackage{listings}
\usepackage{tikz}
\usepackage{bm}
\usepackage{hhline}
\usepackage{enumitem}
\usepackage{soul}
\usepackage{setspace}
\usepackage{minitoc}
\usepackage{tcolorbox}
\usepackage{dashrule}
\usepackage{xspace}
\tcbuselibrary{most}

\newcommand\eg{\emph{e.g.}} 
\newcommand\ie{\emph{i.e.}}

\newlength\savewidth
\newcommand\thline{\noalign{\global\savewidth\arrayrulewidth
  \global\arrayrulewidth 1pt}\hline\noalign{\global\arrayrulewidth\savewidth}}
\newcommand\shline{\noalign{\global\savewidth\arrayrulewidth
  \global\arrayrulewidth 0.75pt}\hline\noalign{\global\arrayrulewidth\savewidth}}

\makeatletter
\newcommand{\customdashline}[3]{%
    \noalign{\vbox{\hrule height 0pt%
        \hbox to\linewidth{\cleaders\hbox to \dimexpr #1 + #2\relax{\hss\rule{#1}{#3}\hss}\hfill}%
    }}%
}
\makeatother

\newcommand{\tablestyle}[2]{\setlength{\tabcolsep}{#1}\renewcommand{\arraystretch}{#2}\centering\scriptsize}
\renewcommand{\paragraph}[1]{\vspace{+.2em}\noindent\textbf{#1}}
\newcolumntype{x}[1]{>{\centering\arraybackslash}p{#1pt}}
\newcolumntype{y}[1]{>{\raggedright\arraybackslash}p{#1pt}}
\newcolumntype{z}[1]{>{\raggedleft\arraybackslash}p{#1pt}}
\newcolumntype{C}{>{\centering\arraybackslash}X}
\newcommand{\app}{\raise.17ex\hbox{$\scriptstyle\sim$}}

\definecolor{deemph}{gray}{0.58}

\definecolor{baselinecolor}{gray}{.9}

\newcommand{\RNum}[1]{\uppercase\expandafter{\romannumeral #1\relax}}

\definecolor{userborder}{rgb}{0.2608, 0.7071, 0.0218}
\definecolor{brightgreen}{rgb}{0.001, 0.921, 0.38}
\definecolor{darkgreen}{rgb}{0.3608, 0.6471, 0.2118}
\definecolor{userfont}{RGB}{0, 0, 0}
\definecolor{darkred}{rgb}{0.8,0.02,0.02}
\definecolor{ddarkred}{rgb}{0.5,0.02,0.02}
\definecolor{highlightred}{HTML}{ffcbc0}
\definecolor{cvprblue}{rgb}{0.21,0.49,0.74}  
\definecolor{cadmiumgreen}{rgb}{0.0, 0.42, 0.24}
\definecolor{aliceblue}{rgb}{0.91, 0.94, 0.97}
\definecolor{darkblue}{rgb}{0.83, 0.89, 0.97}
\definecolor{Blue9}{rgb}{0.098,0.3,0.9}
\definecolor{paviolet}{HTML}{DEDBE9}
\definecolor{pasky}{HTML}{D5EBEE}
\definecolor{payellow}{HTML}{FFF6EA}
\definecolor{pared}{HTML}{F9E5ED}

\newcommand{\cmark}{\ding{51}}%
\newcommand{\xmark}{\ding{55}}%

\newcounter{takeawayonly}

\usepackage{lipsum}
\usepackage{adjustbox}

\usepackage{listings}

\def\figref#1{figure~\ref{#1}}

\def\eqref#1{equation~\ref{#1}}

\def\1{\bm{1}}

\DeclareMathAlphabet{\mathsfit}{\encodingdefault}{\sfdefault}{m}{sl}
\SetMathAlphabet{\mathsfit}{bold}{\encodingdefault}{\sfdefault}{bx}{n}

\title{Gaze Attention: Query-Adaptive Visual Routing for Efficient Multimodal LLMs}

\author[1,2,*]{Junha Song}
\author[1]{Byeongho Heo}
\author[1]{Geonmo Gu}
\author[2]{Jaegul Choo}
\author[1,\dagger]{Dongyoon Han}
\author[1,\dagger]{Sangdoo Yun}

\affiliation[1]{NAVER AI Lab}
\affiliation[2]{KAIST}

\contribution[*]{Project Lead}
\contribution[\dagger]{Corresponding Author}

\date{\today}
\metadata[Project page]{\href{https://june-page.github.io/gaze-attention}{github.io/gaze-attention}}
\metadata[Code]{\href{https://github.com/junha1125/Gaze-Attention}{github.com/Gaze-Attention}}

\hypersetup{
  pdftitle={Gaze Attention: Query-Adaptive Visual Routing for Efficient Multimodal LLMs},
  pdfauthor={Junha Song, Byeongho Heo, Geonmo Gu, Jaegul Choo, Dongyoon Han, Sangdoo Yun}
}

\abstract{
When humans describe a visual scene, they do not process the entire image uniformly; instead, they selectively fixate on regions relevant to their intended description. 
In contrast, current multimodal large language models (MLLMs) attend to all visual tokens, 
leading to diluted focus and unnecessary computational overhead. 
Existing efficiency methods often compress or discard visual information before generation, potentially losing details needed for later predictions.
In this work, we introduce Gaze Attention, a mechanism that enables MLLMs to select visual regions according to the needs of each generation step. 
By grouping visual tokens into spatial regions and selecting those relevant to the current prediction, Gaze Attention reduces attention computation while focusing on relevant visual content. We further introduce learnable context tokens that summarize images or video frames, preserving global context under selective attention.
Experiments on 13 image and 6 video understanding benchmarks demonstrate that Gaze Attention matches or surpasses dense-attention baselines while using up to 90\% fewer visual KV entries. It also achieves higher average performance than KV-cache eviction baselines under matched visual KV budgets.
}

\begin{document}
\maketitle
\vspace{2em}

\section{Introduction}
\label{sec:intro}

Multimodal large language models (MLLMs) encode images or videos into visual embeddings, which are concatenated with textual inputs and processed by an LLM~\citep{llava, llavaonevision, qwen2vl, internvl}.
During generation (\ie, next token prediction), the models attend to all visual tokens, including regions that may be irrelevant to the current prediction.
In contrast, human visual perception is inherently selective: humans shift their gaze to fixate on regions relevant to their intended description~\citep{yarbus, henderson2003human}.
This discrepancy suggests that current MLLMs lack selective visual attention, motivating the development of such mechanisms.

\begin{figure}[t]
    \centering
    \includegraphics[width=1.\linewidth]{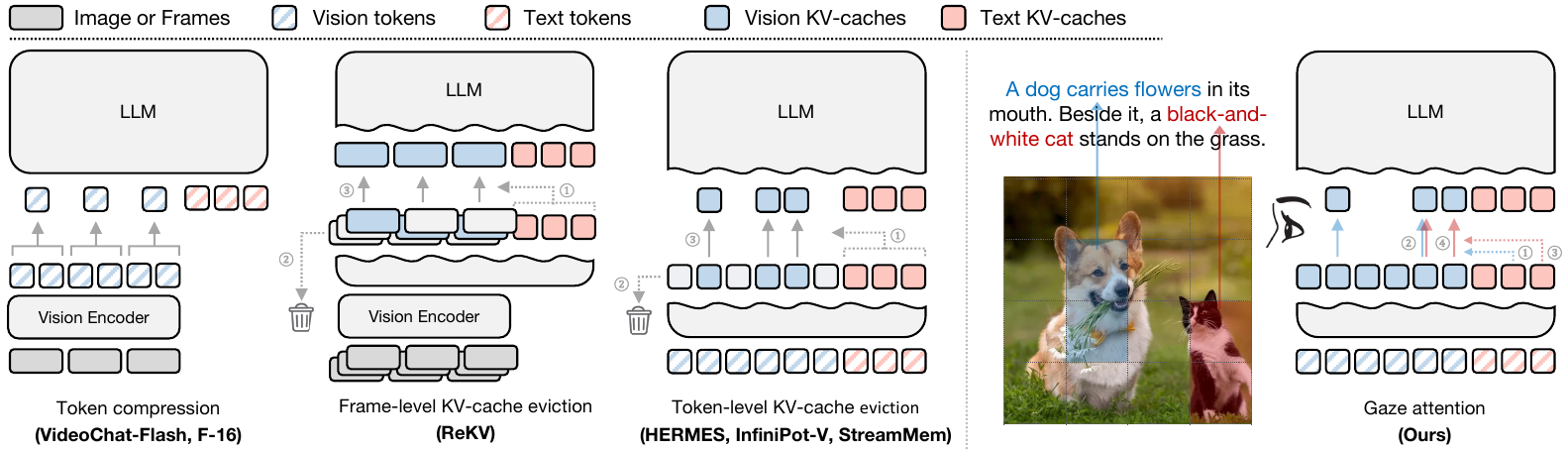}
    \vspace{-1.2em}
    \caption{
    \textbf{Overview of efficient MLLM approaches.}
    Token compression~\citep{videochatflash, f16} and KV-cache eviction~\citep{rekv, hermes, infinipotv, streammem} reduce visual KV cache entries during prefill, before it is known which regions will be needed for subsequent predictions.
    Gaze Attention instead selects and attends to query-relevant gaze regions, allowing the model to focus on what matters at each generation step.
    }
    \label{fig:figure1}
    \vspace{-1em}
\end{figure}

\begin{figure}[t]
    \centering
    \includegraphics[width=1.\linewidth]{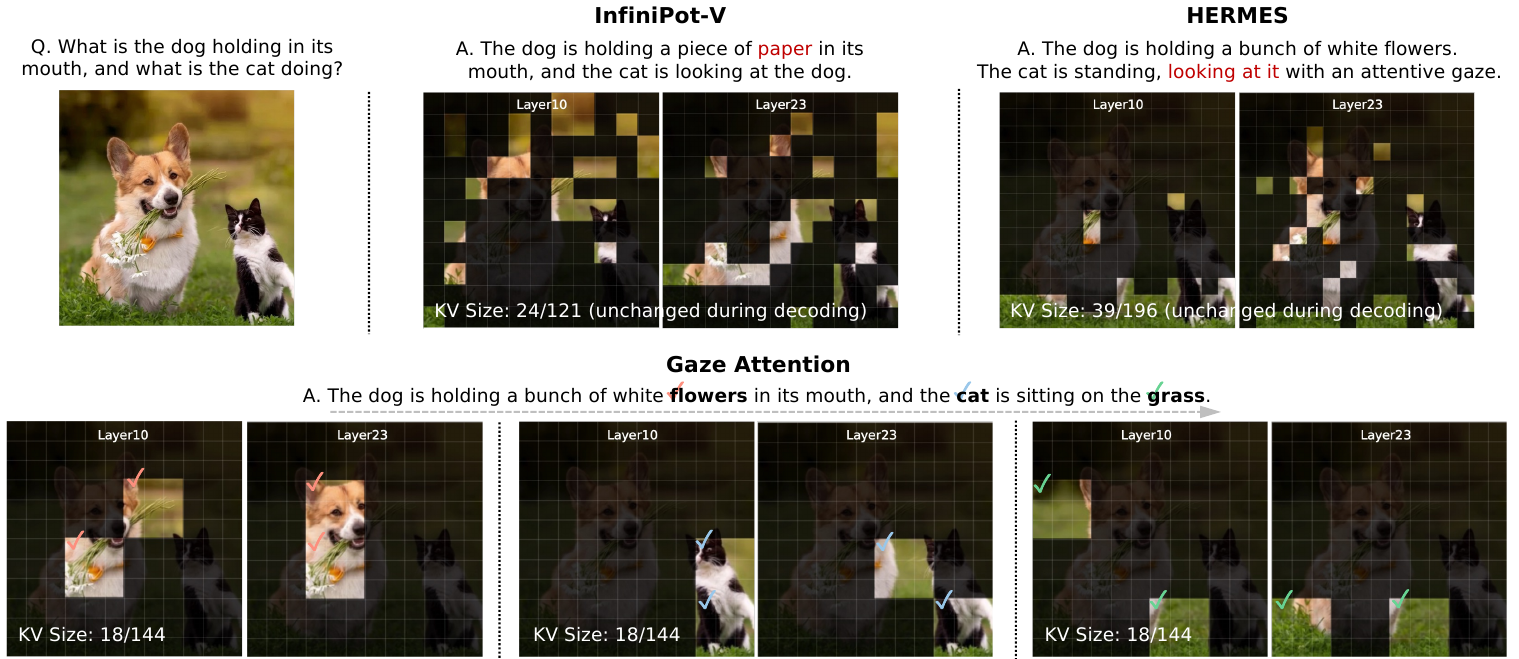}
    \vspace{-1.2em}
    \caption{
    \textbf{Comparison of visual token selection during generation.}
    InfiniPot-V~\citep{infinipotv} and HERMES~\citep{hermes} select visual tokens during prefill and retain the same set throughout decoding, which can omit information needed for later predictions.
    In contrast, Gaze Attention selects regions at each generation step, shifting its focus as the model describes the \textit{flowers}, the \textit{cat}, and the \textit{grass}.
    Highlighted regions show the visual tokens attended to at the indicated layers.
    More examples are provided in \cref{fig:image-gaze-supple}.
    }
    \label{fig:image-gaze-main}
    \vspace{-.8em}
\end{figure}

Existing efforts have explored reducing the visual information that MLLMs attend to, primarily for efficiency rather than selective gaze. These approaches are illustrated in \figref{fig:figure1}.
Some methods compress visual tokens before they are passed to the LLM~\citep{longvlm, videochatflash, f16}, while others evict visual KV cache entries based on question-derived importance scores~\citep{rekv, meda, hermes, streammem, streamkv}.
Both approaches determine the available visual information during the prefill stage, when it is not yet known which regions will matter for each prediction (\cref{fig:image-gaze-main}).
This can be particularly problematic when prediction requires information from multiple distinct regions without omissions or loss of detail.

In this work, we propose Gaze Attention, a mechanism that allows the model to focus on the visual regions needed for each generation step.
Specifically, the cached key vectors of visual tokens are spatially grouped into gaze regions, each represented by a lightweight descriptor. At each decoding step, the model routes attention to a small subset of regions based on these descriptors and attends exclusively to them.
This enables the model to focus on query-relevant regions, improving generation quality while reducing attention computation.
Selective attention alone, however, can limit access to global scene information. To address this, we introduce learnable context tokens that summarize each image or video frame. These tokens remain accessible throughout generation, allowing the model to retain global context as it shifts attention between local regions.
This is similar to how humans take in a scene as a whole before focusing on specific details.

We evaluate Gaze Attention on 13 image and 6 video understanding benchmarks. While attending to up to 90\% fewer visual KV entries, Gaze Attention matches or outperforms dense-attention baselines. It also achieves higher average performance than KV-cache eviction baselines under matched visual KV budgets. Visualization analyses further show that the model shifts its focus toward regions relevant to each prediction, producing more concentrated attention maps than dense attention (\cref{fig:attention-vis}). These results suggest that selective visual attention can preserve task performance while reducing attention computation.

We summarize our contributions as follows:

\begin{itemize}[leftmargin=1.2em, labelsep=0.6em, itemsep=0.0em, topsep=-0.2em]
    \item We identify a limitation of efficient MLLM approaches: pre-selecting visual information can discard fine-grained details and limit adaptation to changing needs during generation.
    \item We introduce Gaze Attention, a visual routing mechanism enabling MLLMs to focus only on what matters for each decoding query.
    \item To mitigate the loss of global context caused by localized attention, we further present learnable context tokens that provide each query with a global visual summary.
    \item We demonstrate that our approach achieves competitive performance under limited visual KV budgets across 13 image and 6 video understanding benchmarks.
\end{itemize}

\section{Related Works}

\paragraph{Multimodal LLMs.}
Recent MLLMs adopt a similar pipeline that encodes images or videos into visual features, and processes them with an LLM backbone~\citep{llava15, llavanext, kimivl, eagle25, molmo2, hoverpivotrl, mmser}.
As images and video frames are information-dense, they typically yield far more visual tokens than text.
This issue becomes more pronounced for high-resolution images~\citep{llavaonevision,  qwen2.5, internvl-2.5} and videos~\citep{cambrian-s, thinking-in-space, streamingvlm}, which require an enormous number of tokens to faithfully represent their content.
Despite the growing demand on LLM backbones to process large amounts of visual information, there remains limited exploration into how such information should be handled effectively within the LLM.

\paragraph{Visual Saliency \& Gaze Modeling.}
Computational modeling of visual attention has a long history in computer vision.
Classical saliency models~\citep{itti1998saliency} predict human fixations from low-level features, while DeepGaze~II~\citep{deepgaze2} and its successors~\citep{deepgaze2e} improve fixation prediction using recognition-network features.
Related efforts study gaze following~\citep{recasens2015gaze} and task-driven scanpath prediction~\citep{chen2021scanpath}.
Beyond predicting where humans look, recent vision-language models incorporate explicit human gaze in several ways.
Voila-A~\citep{voilaa} conditions models on user gaze traces or heatmaps through perceiver modules, while Gaze-VLM~\citep{gazevlm} regularizes transformer attention toward human gaze maps during training.
Other approaches use eye-tracking signals to resolve referential ambiguity in visual question answering~\citep{madinei2026gazevlm} or formulate gaze understanding tasks within a VLM~\citep{mathew2025gazevlm}.
These methods rely on gaze inputs or supervision and largely retain dense visual attention.
In contrast, Gaze Attention requires neither gaze annotations nor gaze input; it learns word-conditioned visual focus shifts through query-adaptive routing and restricts each decoding query to a small subset of visual KV regions.

\paragraph{VideoLLM with Token Reductions.}
To reduce the cost of long visual contexts, prior work has explored compressing visual tokens before they are processed by the LLM. Several strategies include token resampling~\citep{videollama, koala, llamavid, longvlm} and pooling~\citep{videochatgpt, nvila, f16}, which reduce token count by compressing or aggregating visual information. Other works adopt importance-based pruning to remove less relevant tokens~\citep{fastv, videochatflash}. While these methods improve efficiency, they often blur or discard fine-grained details that may be needed later during generation.

\paragraph{VideoLLM with KV-Cache Retrievals.}
Since attention is the dominant computational cost in LLMs, several works aim to improve efficiency, particularly in streaming scenarios, by reducing the amount of visual KV cache~\citep{rekv}.
Some works focus on removing low-importance entries~\citep{meda, hermes, livevlm}, while others consolidate redundant entries within a fixed budget~\citep{rekv, infinipotv, streammem, streamkv}.
In most cases, importance scores between question and visual tokens are computed once at the first generation step, and only high-scoring entries are memorized. 
This limits adaptivity during decoding, especially when later queries require different visual details. 
Our method instead routes each query to relevant visual chunks, maintaining dynamic access throughout generation.

\paragraph{Long Video Generation \& Long-Context LLM.}
Long video generation also faces the prohibitive cost of dense attention. While LCT~\citep{lct} relies on full attention, MoC-VideoGen~\citep{moc-videogen} demonstrates the effectiveness of query-adaptive context selection. In parallel, scaling the context window of LLMs has been widely studied through efficient attention mechanisms such as sliding windows~\citep{mistral, longformer} and sparse attention~\citep{mInference, bigbird}.
Recent works such as NSA~\citep{nsa}, MoBA~\citep{moba}, and DSA~\citep{dsa} show that trainable sparse attention can match dense baselines, and KVzip~\citep{kvzip, fastkvzip} demonstrates effective query-agnostic KV cache compression for long-context text LLMs.
However, these methods target unimodal text and do not address selective attention over visual tokens. Our work bridges this gap by studying selective visual attention for multimodal understanding.

\section{Method} 
\label{sec:method}

\begin{figure}[t]
    \centering
    \includegraphics[width=1.\linewidth]{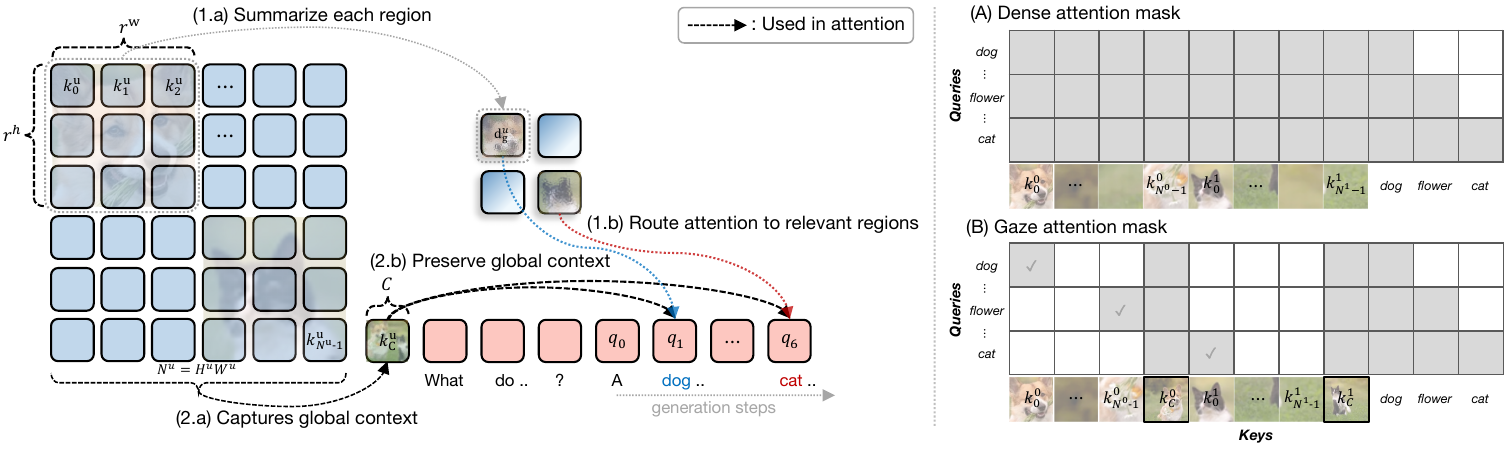}
    \vspace{-1.7em}
    \caption{
    \textbf{Overview of Gaze Attention.}
    Gaze Attention allows MLLMs to
    selectively attend to query-relevant regions.
    Visual keys $k_i^u$ are grouped into gaze regions, each summarized by a lightweight descriptor $d_g^u$.
    Each decoding query $q$ uses these descriptors to select which regions to attend to.
    Learnable context tokens are processed by the LLM to provide global visual context that remains accessible to every query.
    }
    \label{fig:gaze-attention}
    \vspace{-.5em}
\end{figure}

This section presents Gaze Attention and describes its key components: gaze regions, descriptors, and visual routing~(\cref{sec:region-descriptor-routing}), learnable context tokens~(\cref{sec:learnable-context-tokens}), and attention formulation~(\cref{sec:gaze-attention}).
A method overview is illustrated in~\figref{fig:gaze-attention}.

\subsection{Preliminaries: Dense Attention in MLLMs}
\label{sec:prelim}

An MLLM receives visual tokens from a vision encoder, followed by text tokens processed by the LLM. For a visual unit $u$ (\ie, an image or a video frame), let $\{x_i^u\}_{i=0}^{N^u-1}$ denote its visual tokens arranged on an $H^u \times W^u$ grid, where $N^u = H^u W^u$.
As these tokens pass through each LLM layer, they are projected into key--value pairs $\{(k_i^u, v_i^u)\}_{i=0}^{N^u-1}$ and stored in the KV cache $(K, V)$. In dense attention, the query $q_j$ at decoding step $j$ attends to all cache entities:
\vspace{-.2em}
\begin{equation}
  \mathrm{Attn}(q_j, K, V)
  = \mathrm{Softmax}\!\left(\frac{q_j K^\top}{\sqrt{d}}\right) V,
  \label{eq:dense}
\end{equation}
where $d$ is the key dimension.
This makes every visual token accessible, regardless of its relevance to the current prediction. In contrast, humans direct their gaze toward regions relevant to what they are about to describe. Motivated by this, we design a region-based attention mechanism that attends only to the visual regions needed at each generation step.

\subsection{Gaze Regions, Descriptors, and Visual Routing}
\label{sec:region-descriptor-routing}
Visual data, unlike text, is inherently structured in space and time, making one-dimensional partitioning unsuitable for selective attention.
We therefore reshape the visual keys into their original spatial and temporal layout and partition them into regions, each containing $r^h \times r^w \times r^t$ tokens. Let $\mathcal{R}_g^u \subseteq \{0,\ldots,N^u-1\}$ denote the set of token indices belonging to region $g$ in visual unit $u$.
Each region is summarized by a descriptor $d_g^u$, obtained by mean-pooling the key vectors in $\mathcal{R}_g^u$. This descriptor serves as a compact routing unit, reducing the candidate set from the full token sequence to the region level.
At decoding step $j$, the query $q_j$ measures its similarity to the region descriptors $\{d_g^u\}$ and selects the $\mathrm{TopK}$ regions:
\begin{equation}
  s_g^u = q_j^\top d_g^u, \quad \mathcal{G}_j = \mathrm{TopK}(\{s_g^u\}_{u,g}, K),
\end{equation}
where $\mathcal{G}_j$ denotes the selected visual-unit and region index pairs. The query then attends only to the visual key--value pairs within these regions, \textbf{allowing the model to focus its attention computation on query-relevant visual content without being diluted by irrelevant visual tokens}.
Crucially, unlike KV cache eviction methods~\citep{infinipotv,hermes} that discard entries based on a one-time importance estimate, our approach retains the visual KV cache and re-selects regions per query at every decoding step. This means a region ignored at one step can still be attended to at a later step if it becomes relevant.
This enables MLLMs to adapt their visual focus at each decoding step, similar to how humans shift their gaze to examine details while describing different subjects.


\paragraph{Towards Stable Training.}
Training visual routing from scratch can be unstable~\citep{moba, nsa} because the relationship between the descriptors and the decoding queries is weak at the beginning of training. We therefore adopt a progressive $\mathrm{TopK}$ schedule~\citep{gao2024seerattention, moc-videogen}, starting from a dense setting and gradually reducing $\mathrm{K}$ as training proceeds. While the $\mathrm{TopK}$ operation itself is non-differentiable, the selected regions still receive gradient signals through the attention computation, which flow back to the visual keys $k_i^u$ for $(u,g) \in \mathcal{G}_j$ and $i \in \mathcal{R}_g^u$.
This encourages the key projections to produce increasingly discriminative routing signals, leading to more stable routing behavior.

\subsection{Learnable Context Tokens}
\label{sec:learnable-context-tokens}

Localized attention can limit access to global scene information. To alleviate this, we introduce a small set of learnable context tokens. For each visual unit $u$, these tokens are appended to its visual tokens at the input of the LLM. The attention mask restricts each context token to attend only to tokens from its corresponding unit, producing compact summary keys $k_C^u$. This design incurs negligible overhead, as it adds only a few extra token embeddings processed jointly within the LLM's standard parallel computation. Unlike routed visual regions, which vary per query, $k_C^u$ remains visible throughout generation, providing a persistent global summary even when most visual regions are hidden.

\subsection{Gaze Attention}
\label{sec:gaze-attention}

We now combine the above components into a unified attention formulation. For the query $q_j$ at decoding step $j$, the key--value pairs are constructed from three sources: textual tokens, global context keys $k_C^u$, and visual tokens from the routed regions. Specifically, we define
\begin{equation}
K_{\mathrm{Gaze}}(q_j)
=
K_{\mathrm{text}}
\cup
\left\{
 k_C^u
\mid
\forall u
\right\}
\cup
\left\{
k_i^u
\;\middle|\;
(u,g) \in \mathcal{G}_j,\; i \in \mathcal{R}_g^u
\right\},
\label{eq:gaze_k}
\end{equation}
where $V_{\mathrm{Gaze}}(q_j)$ is constructed analogously from the corresponding values of the selected tokens and context tokens. The resulting attention is then given by:
\begin{equation}
  \mathrm{GazeAttn}(q_j)
  =
  \mathrm{Softmax}\!\left(
    \frac{q_j K_{\mathrm{Gaze}}(q_j)^\top}{\sqrt{d}}
  \right)
  V_{\mathrm{Gaze}}(q_j).
  \label{eq:gaze_attn}
\end{equation}
This allows the model to preserve linguistic coherence and global visual awareness while allocating most of its visual computation to regions selected for the current query.

\begin{figure}[t]
    \centering
    \includegraphics[width=1.\linewidth]{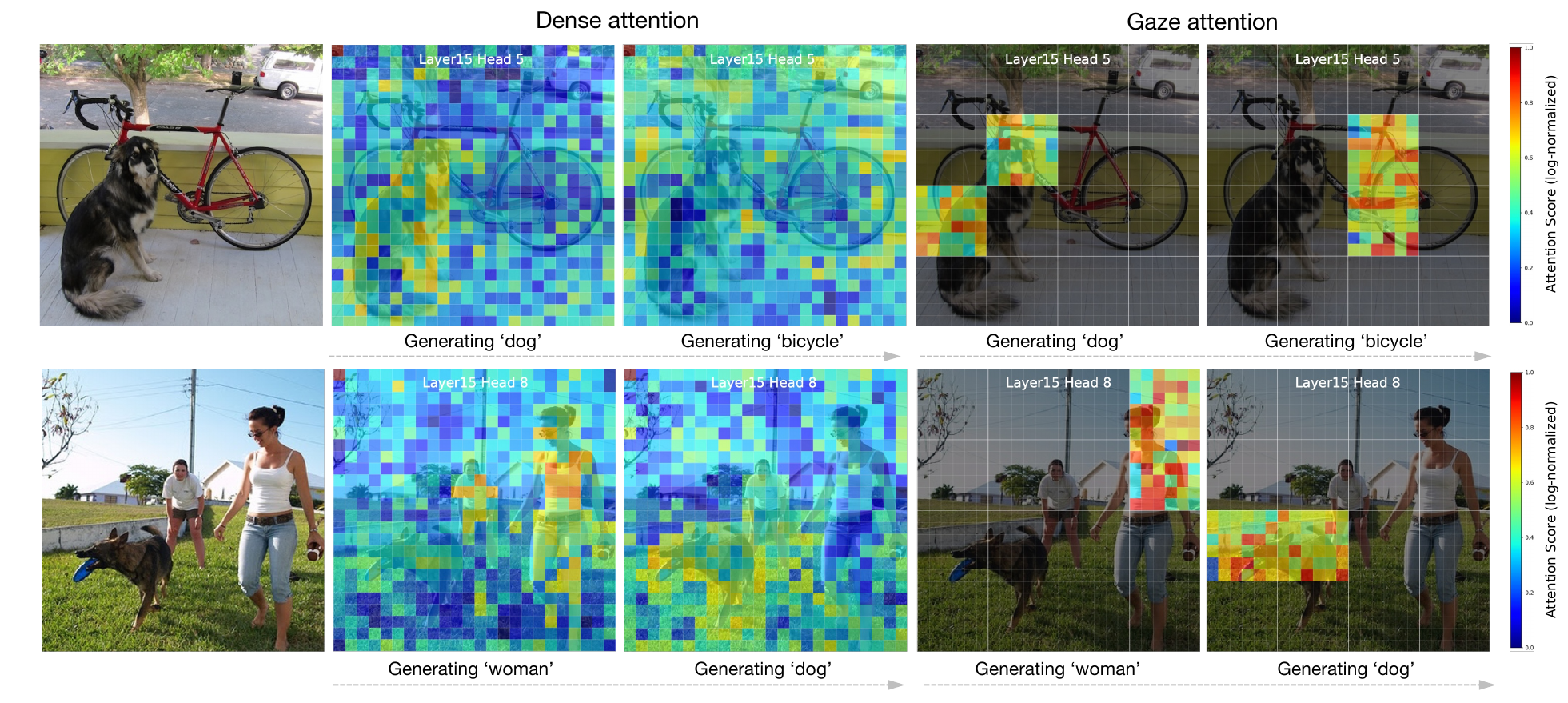}
    \vspace{-2em}
    \caption{
    \textbf{Attention map visualization of dense attention vs.\ gaze attention.} Dense attention spreads across all visual tokens with weak focus on relevant regions. Gaze attention routes over lightweight region descriptors and attends only to the selected regions, yielding more concentrated attention on needed visual content.
    }
    \label{fig:attention-vis}
    \vspace{-.5em}
\end{figure}

\subsection{Cost Analysis.}
Gaze Attention restricts each query to attend only to the selected regions and context tokens, rather than all visual tokens.
As a concrete example, consider an 8-frame video where each frame produces 576 visual tokens after the vision encoder, yielding $N_v{=}4608$ in total. With region size $m = r^h r^w r^t = 36$ ($G{=}128$ regions), $\mathrm{K}{=}20$ selected regions, and $|C|{=}4$, $U{=}8$,
the attended set per query is $\mathrm{K}m + |C|U = 752$, reducing the attention FLOPs and per-head score matrix to roughly $752/4608 \approx 16.3\%$ of dense attention.
For video inputs where $N_v$ can exceed $10^4$, the computational and memory savings become even more substantial. 
We provide detailed FLOPs and memory measurements in \cref{sec:flop}.

\section{Results}

\subsection{Experimental Setup}

\paragraph{Architecture \& Hyperparameters.}
Our implementation is based on the Cambrian codebase~\citep{cambrian}. We pair SigLIP2-So/16 384px~\citep{siglip2} as the vision encoder with either Qwen3-4B~\citep{qwen3} or Qwen2.5-7B~\citep{qwen2.5} as the LLM, denoted Cambrian-4B and Cambrian-7B, respectively. Both configurations use a 2-layer MLP as the multimodal projector.
We optionally adapt a token compression method~\citep{f16} where the projector is augmented with a $3 \times 3$ convolution layer that reduces the number of visual tokens by $4\times$; this is used by default for video inputs.
Gaze regions are set to $r^h \times r^w \times r^t = 6 \times 6 \times 1$ by default, except for image inputs with token compression where we use $3 \times 3 \times 1$. We use $|C| = 4$ learnable context tokens per visual unit.
Following prior work on sparse attention~\citep{infinipotv,moc-videogen,moba},
we allow each attention head to independently select its own set of regions.

\paragraph{Training \& Dataset}
We follow the training strategy of the Cambrian codebase. Our training consists of two stages: in Stage~1, we train only the multimodal projector on the Cambrian-Align dataset~\citep{cambrian} to align the visual and language embedding spaces; in Stage~2, we perform full-parameter fine-tuning. For image training, we use the Cambrian-7M dataset~\citep{cambrian}. For video training, we sample a 20\% subset of the Cambrian-S-3M dataset~\citep{cambrian-s} to keep the training cost manageable; each configuration requires approximately 384 H100 GPU-hours for both image and video training. Additional details are in \cref{sec:training-details}.
Following standard practice in video LLMs~\citep{videollama,llavaonevision}, video frames are extracted at 1 FPS up to a configured maximum, beyond which uniform subsampling is applied.

\paragraph{Evaluation.}
Recent methods in visual token reduction and KV cache compression are built on different architectures and training recipes, making direct comparison difficult. We therefore build baselines under an identical architecture and token-matched training setup to isolate the effect of our method. We additionally report published results of existing methods on their original architectures for reference. More details of the evaluation benchmarks are provided in \cref{sec:evaluation-benchmarks}.

\subsection{Image Question \& Answering}

\paragraph{Attention map visualization.}
To provide an intuitive comparison between Gaze Attention and dense attention, we visualize their attention maps in~\cref{fig:attention-vis}. In this example, dense attention attends over all 576 visual cache entries, whereas Gaze Attention attends to only 72 selected entries (\ie, 2 regions of 36 entries each). Dense attention yields diffuse and ambiguous patterns, as it lacks an explicit mechanism to filter out irrelevant visual content. In contrast, Gaze Attention first selects a small subset of regions, producing more concentrated attention patterns at lower attention cost.

\paragraph{Visual routing examples.}
Gaze Attention allows each query to attend to the visual regions it needs. \Cref{fig:image-gaze-main} illustrates how this region selection evolves over the course of generation: much like humans shift their gaze toward the \textit{flowers}, the \textit{cat}, and the \textit{grass} when describing a scene, the model correspondingly adjusts its attended regions at each step. Notably, this behavior emerges without any explicit supervision over where to look, arising solely from the next-token prediction signal.
In addition, we compare with KV-cache eviction methods such as InfiniPot-V~\citep{infinipotv} and HERMES~\citep{hermes}, which determine retained entries during prefill before the required information is known. Here, they appear to preserve a globally dispersed set of tokens, yet produce some incorrect predictions.

\paragraph{Quantitative results.}
\begin{table}[t]
\centering
\small
\setlength\tabcolsep{2.5pt}
\renewcommand{\arraystretch}{1.5}
\begin{adjustbox}{max width=\textwidth}
\begin{tabular}{l|c|c|cccccccc|ccc|ccc}
\shline
\rowcolor[gray]{0.9}
\multicolumn{1}{l|}{} & \multicolumn{1}{l|}{} & \multicolumn{1}{c|}{} & \multicolumn{8}{c|}{\textbf{General}} & \multicolumn{3}{c|}{\textbf{Science \& Math}} & \multicolumn{3}{c}{\textbf{Chart \& OCR}} \\
\rowcolor[gray]{0.9}
Model & KV Size & Avg &
\multicolumn{1}{c}{\rotatebox{90}{MME$^\text{P}$}} &
\multicolumn{1}{c}{\rotatebox{90}{MMB$^\text{EN}$}} &
\multicolumn{1}{c}{\rotatebox{90}{MMB$^\text{CN}$}} &
\multicolumn{1}{c}{\rotatebox{90}{MMVP}} &
\multicolumn{1}{c}{\rotatebox{90}{RealWorldQA}} &
\multicolumn{1}{c}{\rotatebox{90}{ADE}} &
\multicolumn{1}{c}{\rotatebox{90}{OmniBench}} &
\multicolumn{1}{c|}{\rotatebox{90}{GQA}} &
\multicolumn{1}{c}{\rotatebox{90}{SQA$^\text{I}$}} &
\multicolumn{1}{c}{\rotatebox{90}{MMMU$^\text{V}$}} &
\multicolumn{1}{c|}{\rotatebox{90}{MathVista}} &
\multicolumn{1}{c}{\rotatebox{90}{ChartQA}} &
\multicolumn{1}{c}{\rotatebox{90}{OCRBench}} &
\multicolumn{1}{c}{\rotatebox{90}{TextVQA}} \\
\thline
BLIP-2-7B~\citep{li2023blip2}                    & 576 & -- & 1333.7 & -- & -- & 23.3 & 42.4 & -- & 31.5 & 57.3 & 63.0 & 26.8 & 11.3 & 58.8 & 43.7 & 50.9 \\
SPHINX-7B~\citep{sphinx}                    & -- & -- & 1515.8 & -- & -- & 38.7 & 48.5 & -- & 61.7 & 61.1 & 68.6 & 31.6 & 9.8 & 37.8 & 29.8 & 38.3 \\
LLaVA-Next-LLaMA-3-8B~\citep{llavanext}       & 784 & -- & 1526.0 & 74.0 & -- & 40.0 & 60.1 & -- & 67.7 & 65.3 & 72.9 & 39.6 & 11.8 & 69.2 & 58.1 & 64.7 \\
LLaVA-Next-Qwen2-7B~\citep{llavanext}         & 784 & -- & 1453.3 & 75.2 & -- & 49.3 & 60.0 & -- & 69.7 & 63.6 & 73.5 & 36.1 & 30.0 & 67.1 & 54.2 & 63.6 \\
Cambrian-8B-DINOv2-L~\citep{cambrian}           & 576 & -- & 1366.7 & 65.6 & -- & 34.7 & 54.6 & 58.9 & 57.8 & 63.7 & 68.7 & 36.3 & 35.5 & 18.6 & 4.4 & 47.9 \\
Cambrian-8B-CLIP-L~\citep{cambrian}             & 576 & -- & 1585.3 & 72.7 & -- & 31.3 & 56.2 & 59.0 & 56.2 & 64.0 & 77.4 & 36.1 & 37.3 & 59.4 & 48.0 & 62.4 \\
\shline
Qwen2.5-VL-3B~\citep{qwenvl2.5}                    & 256 & \textbf{63.3} & \textbf{1599.0} & \textbf{77.0} & \textbf{77.4} & \textbf{40.7} & \textbf{59.5} & \textbf{63.2} & \textbf{73.3} & \textbf{60.5} & \textbf{80.5} & \textbf{46.6} & \textbf{23.3} & \textbf{70.9} & \textbf{65.2} & \textbf{68.2} \\
\rowcolor[gray]{0.95}
~~+ InfiniPot-V~\citep{infinipotv}                & 128 & 60.3\,{\scriptsize\textcolor{darkred}{(\textbf{$-$3.0})}} & 1574.1 & 74.7 & 74.1 & 35.3 & 54.1 & 60.2 & 70.0 & 58.5 & \textbf{80.5} & 45.8 & 21.8 & 67.0 & 57.3 & 66.1 \\
\rowcolor[gray]{0.95}
~~+ InfiniPot-V~\citep{infinipotv}                 & 50 & 52.8\,{\scriptsize\textcolor{darkred}{(\textbf{$-$10.5})}} & 1426.5 & 68.5 & 67.6 & 24.7 & 48.1 & 50.7 & 65.5 & 52.5 & 79.0 & 44.2 & 18.0 & 51.0 & 38.4 & 60.2 \\
LLaVA-OV-7B~\citep{llavaonevision}                   & 196 & \textbf{62.4} & \textbf{1522.1} & \textbf{81.0} & \textbf{82.2} & \textbf{54.0} & \textbf{58.6} & \textbf{61.1} & 72.9 & \textbf{62.3} & \textbf{95.6} & 45.1 & \textbf{26.3} & \textbf{63.0} & \textbf{51.1} & 44.2 \\
\rowcolor[gray]{0.95}
~~+ HERMES~\citep{hermes}                     & 100 & 57.9\,{\scriptsize\textcolor{darkred}{(\textbf{$-$4.5})}} & 1452.2 & 79.8 & 80.6 & 45.3 & 49.0 & 58.8 & 73.1 & 57.6 & 90.9 & \textbf{48.0} & 21.0 & 36.8 & 39.5 & \textbf{58.0} \\
\rowcolor[gray]{0.95}
~~+ HERMES~\citep{hermes}                     & 40 & 52.3\,{\scriptsize\textcolor{darkred}{(\textbf{$-$10.1})}} & 1323.4 & 75.2 & 74.8 & 34.0 & 50.2 & 52.8 & \textbf{73.7} & 52.0 & 87.6 & 46.0 & 16.8 & 25.0 & 27.1 & 50.5 \\
\shline
Cambrian-4B                    & 576 & 59.7 & 1580.5 & 74.9 & 73.4 & 45.3 & 59.9 & 58.8 & 70.6 & 56.4 & 79.9 & 46.4 & 17.5 & 60.2 & 49.6 & 64.2 \\
\rowcolor[gray]{0.95}
~~+ Gaze attention                         & 288+4 & 61.1\,{\scriptsize($+$1.4)} & \textbf{1589.8} & \textbf{76.4} & 74.8 & 45.3 & \textbf{61.7} & \textbf{60.8} & \textbf{71.3} & 57.7 & 81.5 & \textbf{47.4} & 17.5 & 62.9 & \textbf{52.6} & 66.0 \\
\rowcolor[gray]{0.95}
~~+ Gaze attention                         & 144+4 & \textbf{61.2}\,{\scriptsize($+$1.5)} & 1561.8 & 76.3 & \textbf{76.3} & \textbf{45.7} & 60.9 & 59.8 & \textbf{71.3} & \textbf{58.2} & 82.0 & 46.3 & 19.1 & 64.4 & 52.4 & \textbf{66.2} \\
\rowcolor[gray]{0.95}
~~+ Gaze attention                         & 72+4 & 60.4\,{\scriptsize($+$0.7)} & 1539.3 & 74.6 & 75.0 & 43.3 & 59.5 & 58.0 & 71.1 & 57.8 & \textbf{82.1} & 46.4 & \textbf{19.7} & \textbf{64.9} & 52.0 & 64.6 \\
\arrayrulecolor{gray!30}\cline{2-17}\arrayrulecolor{black}
Cambrian-4B + TokenCompre. & 144 & 57.8 & 1533.6 & 74.3 & 73.0 & \textbf{47.3} & 55.9 & 58.5 & \textbf{71.3} & 53.0 & \textbf{80.7} & \textbf{45.9} & 17.6 & \textbf{52.6} & 42.0 & 60.1 \\
\rowcolor[gray]{0.95}
~~+ Gaze attention                         & 54+4 & \textbf{59.2}\,{\scriptsize($+$1.4)} & \textbf{1537.0} & \textbf{74.7} & 71.6 & \textbf{47.3} & \textbf{56.5} & 59.7 & 71.0 & \textbf{55.8} & 79.3 & 45.3 & \textbf{31.1} & 52.4 & \textbf{44.7} & \textbf{62.7} \\
\rowcolor[gray]{0.95}
~~+ Gaze attention                         & 36+4 & 58.3\,{\scriptsize($+$0.5)} & 1515.3 & 74.0 & \textbf{73.1} & 40.0 & 54.8 & \textbf{60.0} & 70.1 & 55.4 & 79.5 & 45.7 & 30.7 & 52.2 & 42.8 & 62.3 \\
\shline
\end{tabular}
\end{adjustbox}
\vspace{-.5em}
\caption{
\textbf{Evaluation on Image QA tasks.} We evaluate the effect of Gaze Attention (GA) on image understanding benchmarks. 
InfiniPot-V and HERMES suffer significant performance drops as the KV size decreases, whereas GA maintains or improves over the dense-attention baseline.
``KV Size'' denotes the number of visual KV cache entries attended to during generation. GA controls the KV size by adjusting $\mathrm{K}$, the number of gaze regions. The `+4' in KV Size indicates the number of learnable context tokens.
}
\vspace{-.5em}
\label{tab:image-bench}
\end{table}
Table~\ref{tab:image-bench} summarizes results on image understanding benchmarks. Different MLLMs produce varying numbers of visual tokens depending on their vision encoder, and KV reduction methods further decrease this count (\ie, gray background row). For instance, LLaVA-OV-7B uses SigLIP2-So/14 at 384px, producing 729 visual features; after $2{\times}2$ spatial pooling, 196 tokens enter the LLM. For Gaze Attention, the KV size is determined by the number of attended regions $\mathrm{K}$: given 576 visual tokens partitioned into 16 regions of 36 tokens each, setting $\mathrm{K}{=}2$ means each query attends to $36{\times}2=72$ entries.

We begin by examining existing cache eviction methods. Both HERMES and InfiniPot-V show significant performance drops as the KV size is reduced. This is perhaps unsurprising: 
these methods were primarily designed and validated on video benchmarks, 
where eviction still leaves over 1K visual cache entries for the LLM. In this single-image setting, however, the same aggressive reduction can leave as few as ${\sim}$30 entries, 
providing insufficient visual information to support reliable generation. 
This issue is also compounded by the training-free nature of both methods, which introduces a misalignment---the MLLM was never exposed to such sparse visual inputs during training.
In contrast, Gaze Attention remains robust under comparable or more aggressive compression. Moreover, Gaze Attention operates within the LLM's intermediate layers, allowing it to be combined with input-level token compression: applying TokenCompression~\citep{f16} with Gaze Attention at only 36 KV entries yields performance comparable to dense attention with 144 entries.

\begin{figure}[t]
    \centering
    \includegraphics[width=1.\linewidth]{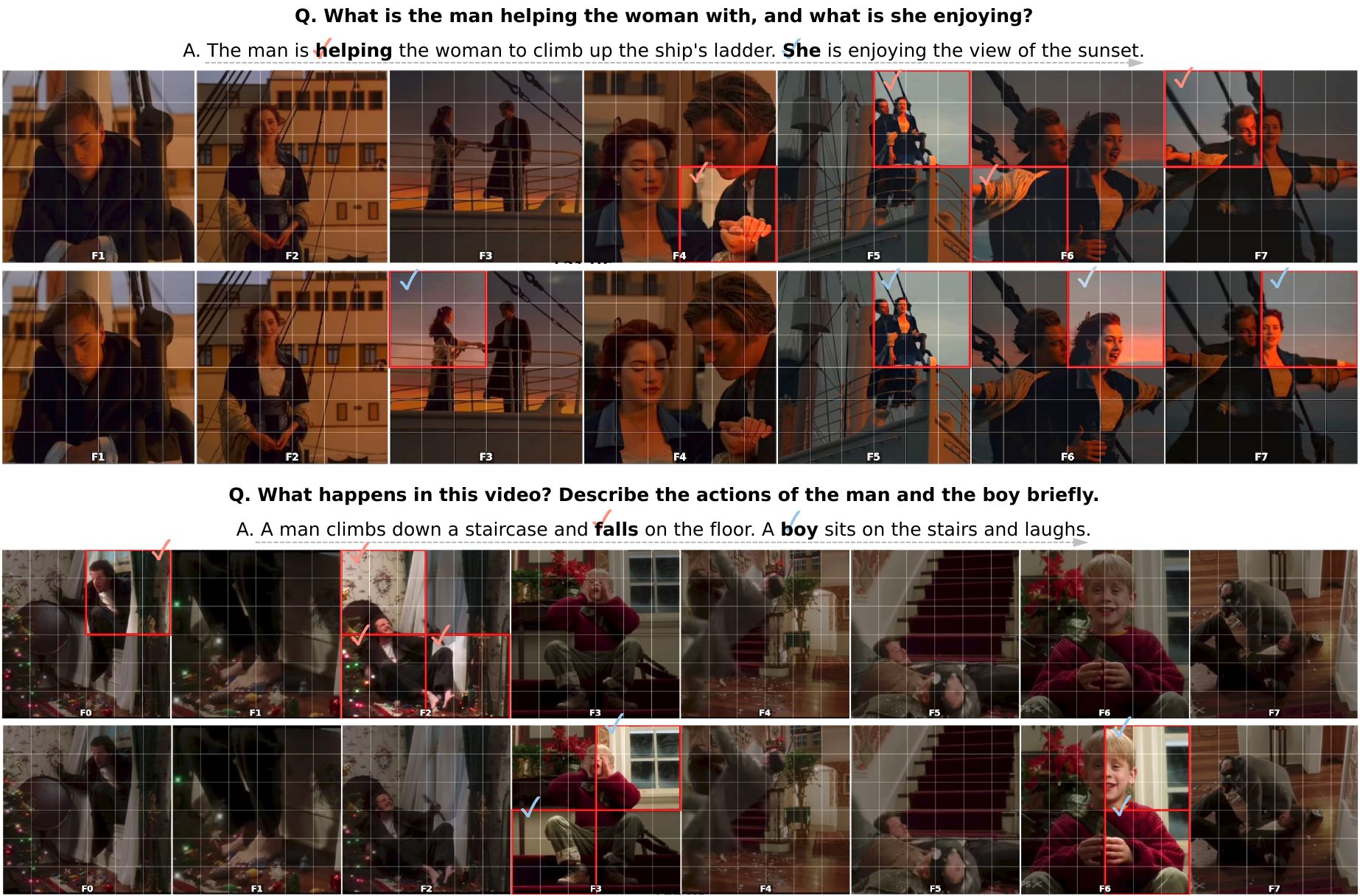}
    \vspace{-1.5em}
    \caption{
    \textbf{Video visual routing examples.} Gaze attention selects relevant spatial regions and temporal frames jointly during generation. When describing different subjects or actions, attended regions (\ie, red boxes) shift to the corresponding frame-region pairs.
    }
    \label{fig:video-gaze-main}
    \vspace{-1.em}
\end{figure}


\subsection{Video Question \& Answering}
The computational burden of visual attention is particularly pronounced in video understanding, where a large number of frames produce an extensive set of visual tokens. We evaluate Gaze Attention on video benchmarks to demonstrate its effectiveness in this setting.

\paragraph{Visual routing examples.}
While the image setting involves only spatial region selection, video understanding additionally requires temporal routing across frames. As shown in \cref{fig:video-gaze-main}, Gaze Attention faithfully performs visual routing in this spatio-temporal setting: different queries select distinct frame-region pairs relevant to the generated content, demonstrating that the gaze shift behavior observed in images extends to the temporal dimension in videos.

\paragraph{Quantitative results.} The results are in \cref{tab:video-bench}. Unlike the image setting, prior eviction methods maintain competitive performance on video benchmarks, consistent with their original design targets. We evaluate Gaze Attention on both Cambrian-4B and Cambrian-7B, and observe that both models improve in average score despite substantial KV size reductions. We hypothesize that selectively attending to relevant regions within lengthy video sequences can be more beneficial, rather than distributing attention uniformly across all frames. Additionally, for Cambrian-4B, a dense-attention baseline constrained to the 4K KV budget averages only 60.6 across the six video benchmarks, whereas Gaze Attention with the same KV budget achieves 63.6. This suggests that Gaze Attention is more effective than simply reducing the number of input tokens under a constrained visual KV budget.

\begin{table}[t]
\centering
\small
\setlength\tabcolsep{4pt}
\renewcommand{\arraystretch}{1.3}
\begin{adjustbox}{max width=\textwidth}
\begin{tabular}{c|l|c|c|cccccc}
\shline
\rowcolor[gray]{0.9}
 & Model & 
\multicolumn{1}{c|}{\rotatebox{0}{KV size}} & 
\multicolumn{1}{c|}{\rotatebox{0}{Gap Avg}} & 
\multicolumn{1}{c}{\rotatebox{0}{VideoMME}} &
\multicolumn{1}{c}{\rotatebox{0}{MLVU}} &
\multicolumn{1}{c}{\rotatebox{0}{EgoSchema}} &
\multicolumn{1}{c}{\rotatebox{0}{LongVideoB.}} &
\multicolumn{1}{c}{\rotatebox{0}{NExT-QA}} &
\multicolumn{1}{c}{\rotatebox{0}{TempCom.}} \\
\shline
& GPT-4o~\citep{gpt4}                & -- & -- & 77.2 & 66.2 & 72.2 & 66.7 & 79.1 & 74.5 \\
& Gemini 2.0 Flash~\citep{gemini2.0} & -- & -- & 70.3 & -- & 71.5 & -- & 81.9 & 76.9 \\
\arrayrulecolor{gray!30}\cline{1-10}\arrayrulecolor{black}
& Qwen2-VL-7B~\citep{qwen2vl}            & 50K & -- & \textbf{63.9} & 65.8 & 65.2 & -- & -- & -- \\
\rowcolor[gray]{0.95}
\cellcolor{white}& ~~+ InfiniPot-V~\citep{infinipotv}     & 6K & $-0.2$ & 62.8\,{\scriptsize($-$1.1)} & 65.8\,{\scriptsize($+$0.0)} & 65.6\,{\scriptsize($+$0.4)} & -- & -- & -- \\
\rowcolor[gray]{0.95}
\cellcolor{white}& ~~+ StreamMem~\citep{streammem}        & 6K & $+0.1$ & 62.1\,{\scriptsize($-$1.8)} & \textbf{65.9}\,{\scriptsize($+$0.1)} & \textbf{67.2}\,{\scriptsize($+$2.0)} & -- & -- & -- \\
\arrayrulecolor{gray!30}\cline{2-10}\arrayrulecolor{black}
& Qwen2.5-VL-3B~\citep{qwenvl2.5}          & 50K & -- & 60.3 & 63.3 & 64.4 & \textbf{59.9} & \textbf{76.8} & \textbf{63.0} \\
\rowcolor[gray]{0.95}
\cellcolor{white}& ~~+ InfiniPot-V~\citep{infinipotv}     & 6K & $-2.1$ & 59.3\,{\scriptsize($-$1.0)} & 62.1\,{\scriptsize($-$1.2)} & 61.8\,{\scriptsize($-$2.6)} & 56.5\,{\scriptsize($-$3.4)} & -- & -- \\
\rowcolor[gray]{0.95}
\cellcolor{white}\multirow{-6}{*}{\rotatebox{90}{Open-weight}} & ~~+ StreamMem~\citep{streammem}        & 6K & $-1.3$ & 59.5\,{\scriptsize($-$0.8)} & 62.3\,{\scriptsize($-$1.0)} & 62.2\,{\scriptsize($-$2.2)} & -- & -- & -- \\
\thline
& InternVL2.5-4B~\citep{internvl-2.5} & -- & -- & 62.3 & -- & 66.6 & 59.3 & 82.5 & 65.2 \\
& PLM-3B~\citep{perceptionlm}         & -- & -- & 54.9 & -- & 66.9 & -- & 83.4 & 69.3 \\
\arrayrulecolor{gray!30}\cline{1-10}\arrayrulecolor{black}
& LLaVA-OV-7B                    & 25K & -- & 58.2 & 64.7 & 60.1 & 57.0 & \textbf{81.0} & \textbf{67.8} \\
\rowcolor[gray]{0.95}
\cellcolor{white}& ~~+ ReKV~\citep{rekv}                        & 6K & $-0.0$ & 57.7\,{\scriptsize($-$0.5)} & -- & \textbf{60.7}\,{\scriptsize($+$0.6)} & 56.8\,{\scriptsize($-$0.2)} & -- & -- \\
\rowcolor[gray]{0.95}
\cellcolor{white}& ~~+ HERMES~\citep{hermes}                      & 6K & $+0.1$ & 58.4\,{\scriptsize($+$0.2)} & -- & 60.2\,{\scriptsize($+$0.1)} & 57.0\,{\scriptsize($+$0.0)} & -- & -- \\
\rowcolor[gray]{0.95}
\cellcolor{white}& ~~+ HERMES~\citep{hermes}                      & 4K & $+0.3$ & 58.9\,{\scriptsize($+$0.7)} & -- & 60.3\,{\scriptsize($+$0.2)} & 56.9\,{\scriptsize($-$0.1)} & -- & -- \\
\arrayrulecolor{gray!30}\cline{2-10}\arrayrulecolor{black}
& Cambrian-4B                                  & 4K & -- & 55.7 & 61.4 & 53.2 & 54.0 & 75.3 & 63.7 \\
& Cambrian-4B                                  & 20K & -- & 58.8 & 65.9 & 53.5 & 57.5 & 77.1 & 60.5 \\
\rowcolor[gray]{0.95}
\cellcolor{white}& ~~+ HERMES~\citep{hermes}                      & 4K & $-0.2$ & 58.1\,{\scriptsize($-$0.7)} & 65.0\,{\scriptsize($-$0.9)} & 53.4\,{\scriptsize($-$0.1)} & 56.2\,{\scriptsize($-$1.3)} & 76.5\,{\scriptsize($-$0.6)} & 62.7\,{\scriptsize($+$2.2)} \\
\rowcolor[gray]{0.95}
\cellcolor{white}& ~~+ Gaze attention                           & 4K & $+$\textbf{1.4} & \textbf{60.4}\,{\scriptsize($+$1.6)} & \textbf{67.5}\,{\scriptsize($+$1.6)} & 53.8\,{\scriptsize($+$0.3)} & 56.8\,{\scriptsize($-$0.7)} & 80.4\,{\scriptsize($+$3.3)} & 62.9\,{\scriptsize($+$2.4)} \\
\rowcolor[gray]{0.95}
\cellcolor{white}& ~~+ Gaze attention                           & 2K & $+1.1$ & 59.4\,{\scriptsize($+$0.6)} & 67.1\,{\scriptsize($+$1.2)} & 53.2\,{\scriptsize($-$0.3)} & 56.6\,{\scriptsize($-$0.9)} & 79.9\,{\scriptsize($+$2.8)} & 63.8\,{\scriptsize($+$3.3)} \\
\arrayrulecolor{gray!30}\cline{2-10}\arrayrulecolor{black}
& Cambrian-7B                                  & 10K & -- & 60.0 & 66.5 & 55.8 & \textbf{58.9} & 78.4 & 60.9 \\
\rowcolor[gray]{0.95}
\cellcolor{white}& ~~+ HERMES~\citep{hermes}                      & 1.5K & $+0.4$ & 60.1\,{\scriptsize($+$0.1)} & 66.3\,{\scriptsize($-$0.2)} & 55.6\,{\scriptsize($-$0.2)} & 58.5\,{\scriptsize($-$0.4)} & 78.1\,{\scriptsize($-$0.3)} & 64.2\,{\scriptsize($+$3.3)} \\
\rowcolor[gray]{0.95}
\cellcolor{white}\multirow{-12}{*}{\rotatebox{90}{Open-weight \& data }} & ~~+ Gaze attention                           & 1.5K & $+1.0$ & \textbf{60.4}\,{\scriptsize($+$0.4)} & 67.0\,{\scriptsize($+$0.5)} & 56.2\,{\scriptsize($+$0.4)} & 58.6\,{\scriptsize($-$0.3)} & 80.4\,{\scriptsize($+$2.0)} & 63.7\,{\scriptsize($+$2.8)} \\
\arrayrulecolor{gray!30}\cline{1-10}\arrayrulecolor{black}
\shline
\end{tabular}
\end{adjustbox}
\vspace{-.5em}
\caption{
\textbf{Evaluation on video QA tasks.} We evaluate Gaze Attention on video understanding benchmarks across two baseline model sizes. It consistently matches or surpasses the dense-attention baseline while attending to a fraction of the visual KV cache. For gaze attention, KV Size includes the learnable context tokens.
}
\vspace{-1.em}
\label{tab:video-bench}
\end{table}

\subsection{Analysis}

\noindent\begin{wrapfigure}{r}{0.45\textwidth}
    \captionsetup{name=Fig., font=small, justification=raggedright, singlelinecheck=false}
    \centering
    \vspace{-1.2em}
    \includegraphics[width=\linewidth]{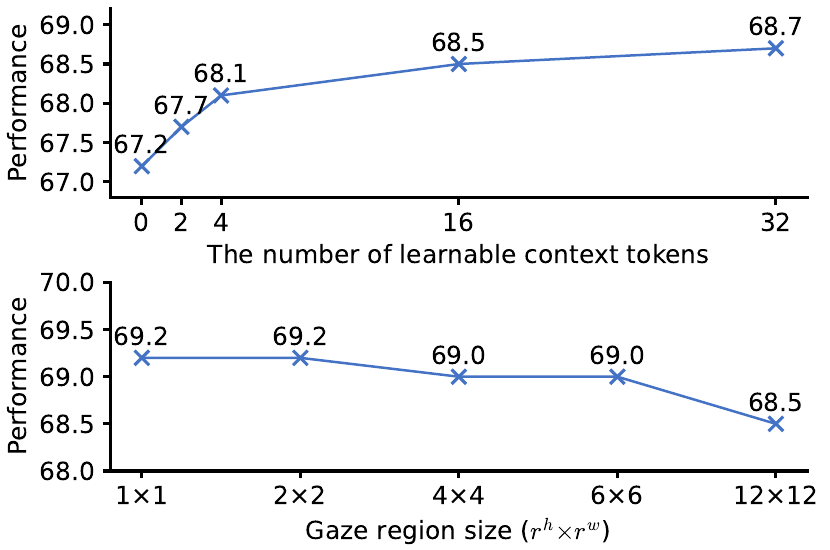}
    \vspace{-2.em}
    \caption{
    \small
    \textbf{Hyperparameter ablation.} Effect of learnable context tokens and gaze region size.
    }
    \label{fig:hyper-ablation}
    \vspace{-1.5em}
\end{wrapfigure}
\paragraph{Number of learnable context tokens.}
We evaluate the effect of learnable context tokens on performance, with results shown in the upper plot of \cref{fig:hyper-ablation}. Removing context tokens entirely ($|C|{=}0$) leads to a notable drop of $-1.5$ compared to $|C|{=}32$, confirming that they provide complementary information beyond the routed regions. Performance improves gradually as the number of context tokens increases; for example, $|C|{=}4$ already yields a $+0.9$ improvement over the no-context baseline. Since context tokens are appended per visual unit, their KV overhead scales with the number of frames in video inputs. Balancing performance and efficiency, we adopt $|C|{=}4$.

\paragraph{Gaze region size.}
Choosing an appropriate region size involves a trade-off: overly large regions include irrelevant tokens, diluting the focus of selective attention, while overly small regions increase the number of routing candidates, undermining efficiency.
The lower plot of \cref{fig:hyper-ablation} confirms this, where finer regions ($1{\times}1$, $2{\times}2$) achieve slightly higher scores but at significantly greater routing cost, while $12{\times}12$ regions are too coarse and lead to a clear performance drop. We adopt $6{\times}6$ as the default, balancing competitive performance with efficient routing.
Fixed-size partitions may misalign with semantic boundaries, motivating content-adaptive partitioning for reliable routing as future work~(\cref{sec:limitations}).

\begin{wraptable}{r}{0.50\textwidth}
\captionsetup{name=Tab., font=small, justification=raggedright, singlelinecheck=false}
\vspace{-.8em}
\centering
\scriptsize
\setlength\tabcolsep{4pt}
\renewcommand{\arraystretch}{1.2}
\begin{tabularx}{\linewidth}{l|c|CCC|c}
\shline
\rowcolor[gray]{0.9}
All & Vision & \multicolumn{3}{c|}{FLOPs (GFLOPs)} & Mem.\ (GB) \\
\rowcolor[gray]{0.9}
attn layers & KV size & Attn & Route & Lct & KV Cache \\
\shline
Dense & 10K & 5.52 & -- & -- & 1.37 \\
Gaze  & 1.5K & 0.89 & 0.19 & 0.15 & 0.27 \\
\arrayrulecolor{gray!30}\cline{1-6}\arrayrulecolor{black}
\rowcolor[gray]{0.95}
Saving & 81.6\% & 83.9\% & -- & -- & 79.8\% \\
\shline
\end{tabularx}
\vspace{-.7em}
\caption{\textbf{Computational cost.} Route and Lct mean the overhead from routing and learnable context tokens.}
\label{tab:cost}
\vspace{-1.2em}
\end{wraptable}

\paragraph{FLOPs \& Memory Usage.}
\label{sec:flop}
\Cref{tab:cost} summarizes the attention FLOPs and KV-cache memory of Gaze Attention versus dense attention. For a fair comparison, all FLOPs are measured without kernel-level fused operators in either method. 
In addition, we adopt the KV-cache offloading strategy of ReKV~\citep{rekv, livevlm}, where the full cache resides in CPU RAM/SSD and only the selected regions are loaded onto the GPU. Under these conditions, Gaze Attention reduces attention FLOPs by approximately 83.9\% and KV-cache memory by 79.8\%. For a matched latency comparison, 
we implement all attention methods using plain PyTorch tensor operations, without built-in attention APIs or fused CUDA kernels such as FlashAttention~\citep{flashattention}. Under this setting, InfiniPot-V and HERMES require $1.4$--$1.7{\times}$ and $1.8{\times}$ the wall-clock time of their respective dense baselines, while Gaze Attention incurs $1.5$--$1.6{\times}$ across image and video inference. This overhead arises from KV selection and gathering before attention, reflecting an engineering challenge shared by existing approaches that we consider an important direction for future effort.

\begin{wraptable}{r}{0.35\textwidth}
\captionsetup{name=Tab., font=small, justification=raggedright, singlelinecheck=false}
\vspace{-.8em}
\centering
\scriptsize
\setlength\tabcolsep{5pt}
\renewcommand{\arraystretch}{1.2}
\begin{tabular}{y{50}|x{50}}
\shline
\rowcolor[gray]{0.9}
Routing & Accuracy (\%) \\
\shline
Random & 11.16$\pm$1.4 \\
Layer 9 & 56.6$\pm$2.3 \\
Layer 15 & \textbf{57.7}$\pm$2.1 \\
Layer 21 & 51.3$\pm$2.7 \\
\shline
\end{tabular}
\vspace{-.6em}
\caption{\textbf{Routing acc.} Results are mean $\pm$ standard deviation over three seeds.}
\label{tab:routing-quality}
\vspace{-1em}
\end{wraptable}

\paragraph{Routing quality.}
We further quantify whether the routed regions correspond to the content being generated. To this end, We sample 200 images from Cambrian-7M and generate a caption for each; \mbox{GPT-5.4~mini} then decomposes the caption into phrases and aligns each generated token to one phrase (\eg, ``a dog carries a flower in its mouth'' is decomposed into \textit{dog}, \textit{flower}, and \textit{mouth}), and SAM3 produces a segmentation mask for each phrase. A token is counted as correct when its routed region overlaps with the predicted mask of its aligned phrase.
As shown in \cref{tab:routing-quality}, Gaze Attention achieves 51.3--57.7\% routing accuracy across the evaluated layers, far exceeding the 11.16\% of random routing, even though the model is trained without any explicit supervision over where to look. Accuracy slightly decreases at layer 21, suggesting that explicit region selection matters more in the earlier and middle layers. Although these automatic proxies introduce noise into the absolute accuracy, the consistent margin over random routing confirms that visual routing selects semantically relevant regions.

\paragraph{Progressive $\mathrm{TopK}$ schedule.}
We ablate the progressive schedule using Cambrian-4B + Gaze Attention with 36+4 visual KV entries. Removing the schedule reduces the average image-benchmark score from 66.4 to 63.5 (\cref{tab:image-bench}). Because routing is learned solely through next-token prediction, query--descriptor alignment is weak early in training. Enforcing sparse routing from the outset forces the model to make hard region selections before reliable routing signals emerge. Starting dense and gradually reducing $\mathrm{K}$ therefore provides more stable optimization and yields the observed gain.

\section{Conclusion}
We presented Gaze Attention, a selective attention mechanism that enables MLLMs to focus on query-relevant visual regions, similar to how humans shift their gaze while describing a scene.
Cached visual keys are spatially grouped into gaze regions, and each decoding query is routed to a small subset of relevant regions.
Learnable context tokens complement this localized attention by preserving global visual context.
Experiments on image and video understanding benchmarks show that our approach matches or outperforms dense-attention baselines while attending to up to 90\% fewer visual KV entries.
It also achieves higher average performance than KV-cache eviction baselines under matched visual KV budgets.
Visualization analyses further show that the model shifts its focus toward regions relevant to each prediction without explicit gaze supervision.

\section*{Acknowledgments}
We gratefully acknowledge NAVER AI Lab for providing the resources and support that facilitated this research. Additional support was provided by the National Research Foundation of Korea (NRF) through a grant funded by the Korean government (MSIT) (No. 2022R1A5A7083908), as well as by the National Supercomputing Center, which provided supercomputing resources and technical support (KSC-2025-CRE-0482).

\bibliography{CITE}
\bibliographystyle{unsrtnat}

\newpage
\renewcommand \thepart{}
\renewcommand \partname{}
\appendix

\renewcommand\thefigure{\Alph{figure}}    
\setcounter{figure}{0}  
\renewcommand\theHfigure{appendix.\Alph{figure}}
\renewcommand\thetable{\Alph{table}}
\setcounter{table}{0} 
\renewcommand\theHtable{appendix.\Alph{table}}

\clearpage



\begin{table}[t]
\centering
\small
\setlength\tabcolsep{3.2pt}
\renewcommand{\arraystretch}{1.25}
\begin{adjustbox}{max width=\textwidth}
\begin{tabular}{y{150}|x{45}|x{75}|x{90}|x{80}|x{50}}
\shline
\rowcolor[gray]{0.9}
Method & Venue & Task & \shortstack{Sparse/Selective\\Attention} & \shortstack{Selection Stage} & Condition \\
\shline
Token reduction~\citep{videollama,koala,llamavid,longvlm,videochatgpt,nvila,f16,fastv,videochatflash} & -- & Multimodal und. & \xmark & Prefill & -- \\
KV eviction~\citep{rekv,meda,livevlm,streamkv} & -- & Multimodal und. & \cmark & Prefill & Question \\
MMInference~\citep{mminference} & ICML'25 & Multimodal und. & \cmark & Prefill & Question \\
VideoNSA~\citep{videonsa} & ICLR'26 & Multimodal und. & \cmark & Prefill & Vision input \\
StreamMem~\citep{streammem} & arXiv'25 & Multimodal und. & \cmark & Prefill & Question \\
InfiniPot-V~\citep{infinipotv} & NeurIPS'25 & Multimodal und. & \cmark & Prefill & Question \\
HERMES~\citep{hermes} & ACL'26 & Multimodal und. & \cmark & Prefill & Question \\
\rowcolor[gray]{0.95}
Gaze Attention & -- & Multimodal und. & \cmark & \textbf{Decoding} & \textbf{Query} \\
\shline
\end{tabular}
\end{adjustbox}
\vspace{-.7em}
\caption{
\textbf{Comparison of methods for efficient multimodal understanding.}
We compare methods by their use of sparse or selective attention, selection stage, and conditioning signal.
Existing approaches focus on prefill, whereas Gaze Attention selects query-relevant visual regions at each decoding step.
}
\vspace{-.5em}
\label{tab:selective-multimodal}
\end{table}

\begin{table}[t]
\centering
\small
\setlength\tabcolsep{3.2pt}
\renewcommand{\arraystretch}{1.25}
\begin{adjustbox}{max width=\textwidth}
\begin{tabular}{y{150}|x{45}|x{75}|x{90}|x{80}|x{50}}
\shline
\rowcolor[gray]{0.9}
Method & Venue & Task & \shortstack{Sparse/Selective\\Attention} & \shortstack{Selection/\\Stage} & Condition \\
\shline
MoC-VideoGen~\citep{moc-videogen} & ICLR'26 & Video generation & \cmark & Decoding & Query \\
KVzip~\citep{kvzip} & NeurIPS'25 & Text und. & \xmark & Prefill & -- \\
Fast KVzip~\citep{fastkvzip} & arXiv'26 & Text und. & \xmark & Prefill & -- \\
MInference~\citep{mInference} & NeurIPS'24 & Text und. & \cmark & Prefill & Question \\
Tri-shape~\citep{scbench} & ICLR'25 & Text und. & \cmark & Prefill & Question \\
BigBird~\citep{bigbird} & NeurIPS'20 & Text und. & \cmark & Prefill & Role-based \\
NSA~\citep{nsa} & ACL'25 & Text und. & \cmark & Decoding & Query \\
MoBA~\citep{moba} & NeurIPS'25 & Text und. & \cmark & Decoding & Query \\
DSA~\citep{dsa} & arXiv'25 & Text und. & \cmark & Decoding & Query \\
\rowcolor[gray]{0.95}
Gaze Attention & -- & \textbf{Multimodal und.} & \cmark & Decoding & Query \\
\shline
\end{tabular}
\end{adjustbox}
\vspace{-.7em}
\caption{
\textbf{Comparison of selective and sparse attention methods across tasks.}
Several methods also use query-conditioned selection during decoding, but target video generation or text understanding.
Gaze Attention instead selects visual regions for multimodal understanding.
We include these methods as related approaches rather than direct baselines.
}
\vspace{-1.em}
\label{tab:selective-other-tasks}
\end{table}

\section{Additional Related Works}
\label{sec:additional-related-works}
\paragraph{Classical Local Attention.}
The idea of focusing on relevant visual evidence has appeared across attention paradigms.
In image captioning, Show, Attend and Tell~\citep{show-attend-tell} introduced word-conditioned spatial attention, and DVSA~\citep{dvsa} learned region--language alignments for grounded descriptions.
Bottom-Up and Top-Down Attention~\citep{bottom-up-top-down} shifted this focus to detected object regions, while LinkNet~\citep{linknet} incorporated object-level relational embeddings for scene-graph reasoning.
Despite differences in architecture and attention design, these methods learn local focus within task-specific captioning or scene-graph pipelines.
Gaze Attention instead targets decoding-time efficiency in MLLMs: it preselects descriptor-scored visual regions for each query and computes attention only over the selected visual KV entries.

\paragraph{Crop--Zoom--Refit Methods.}
Several recent methods enable MLLMs to capture fine-grained image details through a crop--zoom--refit pipeline, with the common goal of improving performance by exposing the model to additional visual evidence.
For example, MLLMs Know Where to Look~\citep{mllms-know-where} uses dense attention maps to localize relevant regions and feeds cropped views back into the model.
Other works introduce explicit region localization into visual reasoning, often through bounding-box prediction or visual operations~\citep{argus,vgr,pixel-reasoner,thinking-with-images}, while VisionThink~\citep{visionthink} decides when a higher-resolution image should be processed.
Gaze Attention differs in both purpose and mechanism: it reduces the accessed visual KV entries for memory efficiency and selects regions before attention via query--region descriptor similarity, avoiding dense attention over all visual tokens.
Accordingly, crop--zoom--refit methods are preferable when fine-grained perception is the primary objective, whereas Gaze Attention is designed for memory-limited on-device settings.
Since these goals are orthogonal, combining the two approaches is an interesting direction for future work.

\section{Comparison with Selective and Sparse Attention}
\label{sec:selective-sparse-comparison}
We organize selective and sparse attention methods by target task, selection stage, and conditioning signal. Our comparison set focuses on methods that reduce visual information for MLLMs.

\Cref{tab:selective-multimodal} summarizes multimodal methods. Token reduction compresses visual tokens before the LLM, while KV-cache methods select or compress entries during prefill. MMInference~\citep{mminference} targets sparse prefill, and VideoNSA~\citep{videonsa} conditions sparse visual attention on the vision input. Gaze Attention instead routes each decoding query to relevant visual regions, enabling word-conditioned focus shifts as illustrated in \cref{fig:figure1}.
Accordingly, we compare primarily with multimodal methods. Because architectures and training recipes differ, we use identical architecture and token-matched training settings where possible and otherwise report published results.
\Cref{tab:selective-other-tasks} lists related learned or query-conditioned methods. The closest decoding-time candidates target video generation or unimodal text understanding and select tokens within a single modality, so we treat them as related methods rather than direct baselines.

\section{Limitations and Future Directions.}
\label{sec:limitations}

Gaze Attention demonstrates that MLLMs can achieve competitive performance while selectively attending to a small fraction of visual tokens, yielding interpretable gaze patterns and substantial KV cache reductions. Nevertheless, the current design has some limitations, which we outline below.

\begin{enumerate}[leftmargin=15pt, itemsep=4pt, topsep=0pt, parsep=0pt]
    \item \textbf{Fixed-size regions.} The current design partitions visual tokens into fixed-size spatial regions, which may not align well with the semantic boundaries of objects or scenes. Content-adaptive region partitioning could yield more reliable gaze routing. In addition, we use a fixed $\mathrm{TopK}$ budget as a simple design choice to demonstrate the feasibility of query-dependent visual routing. Allowing the number of selected regions to adapt to each query or visual input is an important future direction.
    \item \textbf{Descriptor design.} Region descriptors are obtained by straightforwardly mean-pooling the visual cache keys within each region, which avoids additional learnable parameters. However, this treats all tokens within a region equally, including redundant or less informative ones. For instance, if a region contains a salient object, a descriptor more oriented toward that object could help decoding queries select the region more reliably. One possible remedy is to incorporate entropy-based importance weighting~\citep{apt-vit, Adavit} at the cost of extra computation, and finding the right trade-off between descriptor quality and efficiency is worth further investigation.
    \item \textbf{Efficiency--performance trade-off.} Our results primarily demonstrate a favorable memory efficiency--performance trade-off rather than substantial gains in absolute accuracy. Further improving performance under small visual KV budgets remains an important direction.
    \item \textbf{Gaze supervision.} Our model is trained solely with the cross-entropy loss (next-token prediction), without any explicit signal indicating where each decoding query should look. An interesting future direction is to investigate the effect of explicit gaze guidance on visual routing. Consider how a parent reads a picture book to a child: the parent points to specific parts of the illustration while explaining the story, and the child naturally learns to associate the description with the corresponding scene. Likewise, providing the model with such pointing signals during training—indicating which regions to focus on for a given description—could help it learn more accurate vision-language grounding. In practice, this could be realized by combining sentence parsing (\eg, Stanza~\citep{stanza}) with an open-vocabulary detector (\eg, Grounding DINO~\citep{groundingdino}) to automatically generate region-level supervision during training.
    \item \textbf{Routing bias.} Visual routing is learned only from the next-token prediction signal, so it may inherit biases, such as weak positional grounding~\citep{blind-to-position} or socially biased associations~\citep{interpretable-debias}. Measuring whether Gaze Attention amplifies such biases is left to future work.
\end{enumerate}

We believe that addressing these limitations will bring MLLMs closer to perceiving and reasoning about visual scenes in a manner that mirrors human gaze behavior.

\section{Additional Results}
\label{sec:additional-results}

\subsection{Evidence Scale and Spatial Distribution.}
\label{sec:evidence-scale}
Existing MLLM benchmarks do not provide object-size splits. We therefore use counting and spatial reasoning as proxies for large or spatially distributed evidence, fine-grained high-resolution perception as a proxy for small-object evidence, and localization and spatial-relation tasks as proxies for evidence spanning region boundaries.
\Cref{tab:additional-counting,tab:additional-fine-grained,tab:additional-spatial} report the results at 384px and 512px. With 87\% fewer visual KV entries, Gaze Attention remains close to dense attention on counting at 384px (49.48 vs.\ 49.67 normalized average) and surpasses it at 512px (51.87 vs.\ 50.80). It also matches or improves the normalized average on fine-grained perception (24.46 vs.\ 24.17 and 25.95 vs.\ 25.38) and localization and spatial reasoning (48.94 vs.\ 48.25 and 54.17 vs.\ 50.42). These results show that Gaze Attention remains effective for distributed, small, and boundary-crossing evidence under these proxy settings.

\begin{table}[t]
\centering
\scriptsize
\setlength\tabcolsep{1.8pt}
\renewcommand{\arraystretch}{1.25}
\begin{adjustbox}{max width=\textwidth}
\begin{tabular}{y{48}|x{50}|x{45}|x{41} x{44} x{41} x{41} x{41} x{41} x{41}x{41}}
\shline
\rowcolor[gray]{0.9}
Model & ViT & KV size & \shortstack{Avg} & \shortstack{CountBench\\Acc.} & \shortstack{BLINK\\Counting} & \shortstack{MME\\Count} & \shortstack{MMStar\\Obj. Count} & \shortstack{MME-RW\\Vehicle} & \shortstack{MME-RW\\Person} & \shortstack{MME-RW\\Remote} \\
\shline
Dense & So/16-384px & 576 & \textbf{49.67} & \textbf{74.95} & \textbf{62.50} & 175.0 & \textbf{46.74} & 50.00 & 22.00 & 4.00 \\
\rowcolor[gray]{0.95}
Gaze Attention & So/16-384px & 76 (87\%$\downarrow$) & 49.48 & 74.54 & 60.83 & 175.0 & 43.48 & \textbf{52.00} & \textbf{24.00} & 4.00 \\
Dense & So/16-512px & 1024 & 50.80 & 76.58 & 64.17 & \textbf{170.0} & 47.83 & 46.00 & 32.00 & 4.00 \\
\rowcolor[gray]{0.95}
Gaze Attention & So/16-512px & 132 (87\%$\downarrow$) & \textbf{51.87} & \textbf{76.78} & 64.17 & 168.3 & \textbf{50.00} & \textbf{50.00} & 32.00 & \textbf{6.00} \\
\shline
\end{tabular}
\end{adjustbox}
\vspace{-.7em}
\caption{
\textbf{Counting as a proxy for distributed visual evidence.}
With 87\% fewer attended visual KV entries, Gaze Attention achieves a similar average score to dense attention at 384px and a higher score at 512px.
}
\label{tab:additional-counting}
\vspace{-.8em}
\end{table}

\begin{table}[t]
\centering
\scriptsize
\setlength\tabcolsep{1.8pt}
\renewcommand{\arraystretch}{1.25}
\begin{adjustbox}{max width=\textwidth}
\begin{tabular}{y{48}|x{50}|x{45} |x{40} x{40} x{40} x{40} x{40} x{40} x{40}}
\shline
\rowcolor[gray]{0.9}
Model & ViT & KV size & \shortstack{Avg} & \shortstack{MME-RW\\Overall} & \shortstack{MME-RW\\Perception} & \shortstack{MME-RW\\Monitoring} & \shortstack{MME-RW\\Driving} & \shortstack{MME-RW\\Remote} & \shortstack{MME-RW\\Vehicle} \\
\shline
Dense & So/16-384px & 576 & 24.17 & 29.44 & 29.08 & \textbf{23.03} & 29.47 & \textbf{22.00} & 12.00 \\
\rowcolor[gray]{0.95}
Gaze Attention & So/16-384px & 76 (87\%$\downarrow$) & \textbf{24.46} & \textbf{30.54} & \textbf{30.37} & 21.32 & \textbf{29.87} & 20.67 & \textbf{14.00} \\
Dense & So/16-512px & 1024 & 25.38 & 29.08 & \textbf{32.08} & \textbf{24.73} & 32.40 & 22.00 & 12.00 \\
\rowcolor[gray]{0.95}
Gaze Attention & So/16-512px & 132 (87\%$\downarrow$) & \textbf{25.95} & \textbf{30.12} & 31.48 & 23.03 & 32.40 & \textbf{26.67} & 12.00 \\
\shline
\end{tabular}
\end{adjustbox}
\vspace{-.7em}
\caption{
\textbf{Fine-grained perception as a proxy for small-object evidence.}
We evaluate dense attention and Gaze Attention on high-resolution MME-RealWorld tasks at 384px and 512px.
Gaze Attention achieves a higher average score at both resolutions while attending to 87\% fewer visual KV entries.
}
\label{tab:additional-fine-grained}
\vspace{-.8em}
\end{table}

\begin{table}[t]
\centering
\scriptsize
\setlength\tabcolsep{1.8pt}
\renewcommand{\arraystretch}{1.25}
\begin{adjustbox}{max width=\textwidth}
\begin{tabular}{y{48}|x{50}|x{45}|x{40}x{42}x{40}x{40}x{44}x{40}x{40}x{40}}
\shline
\rowcolor[gray]{0.9}
Model & ViT & KV size & \shortstack{Avg} & \shortstack{BLINK\\Spatial Rel.} & \shortstack{BLINK\\Obj. Loc.} & \shortstack{MME\\Position} & \shortstack{MMStar\\Localization} & \shortstack{MMSI\\Obj.-Reg.} & \shortstack{MMSI\\Cam.-Reg.} & \shortstack{MMSI\\Reg.-Reg.} \\
\shline
Dense & So/16-384px & 576 & 48.25 & \textbf{77.62} & 55.74 & 140.00 & 52.50 & 25.88 & 31.33 & \textbf{24.69} \\
\rowcolor[gray]{0.95}
Gaze Attention & So/16-384px & 76 (87\%$\downarrow$) & \textbf{48.94} & 76.22 & 55.74 & \textbf{143.33} & 52.50 & \textbf{32.94} & 31.33 & 22.22 \\
Dense & So/16-512px & 1024 & 50.42 & 78.32 & 56.56 & \textbf{155.00} & 55.00 & 25.88 & 32.53 & 27.16 \\
\rowcolor[gray]{0.95}
Gaze Attention & So/16-512px & 132 (87\%$\downarrow$) & \textbf{54.17} & \textbf{82.52} & 56.56 & 153.33 & \textbf{60.00} & \textbf{36.47} & \textbf{37.35} & \textbf{29.63} \\
\shline
\end{tabular}
\end{adjustbox}
\vspace{-.7em}
\caption{
\textbf{Localization and spatial reasoning as proxies for evidence spanning multiple regions or region boundaries.}
We compare dense attention and Gaze Attention on localization and spatial-relation tasks at 384px and 512px.
Gaze Attention improves the average score while attending to 87\% fewer visual KV entries.
}
\label{tab:additional-spatial}
\end{table}

\subsection{Visual Routing Visualization.}
\label{sec:visual-routing-visualization}
\begin{figure}[t]
    \centering
    \includegraphics[width=0.8\linewidth]{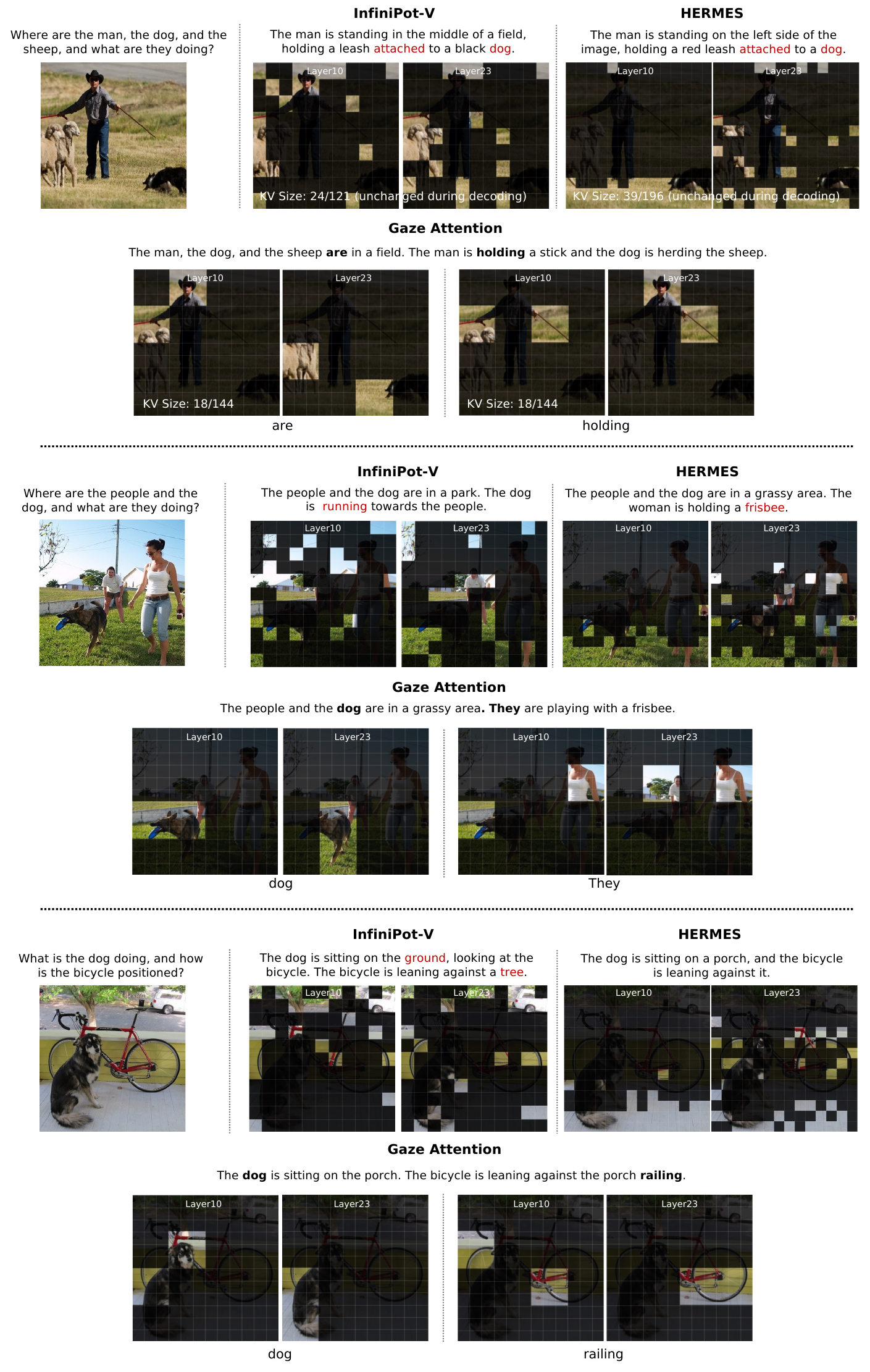}
    \caption{
    \textbf{Additional image visual routing examples.}
    Following the setup in \cref{fig:image-gaze-main}, we compare visual token selection in InfiniPot-V, HERMES, and Gaze Attention.
    The baselines retain the same visual tokens throughout decoding, whereas Gaze Attention selects regions for each generated token.
    Highlighted regions show the attended visual tokens at layers 10 and 23, and red text marks errors in the generated descriptions.
    }
    \label{fig:image-gaze-supple}
    \vspace{-.5em}
\end{figure}


\Cref{fig:image-gaze-supple} provides additional comparisons with InfiniPot-V and HERMES. Unlike their fixed visual cache entries, Gaze Attention changes the routed regions across generated tokens and focuses on evidence relevant to each prediction.

\section{Training Details}
\label{sec:training-details}
We build our models using the LLaVA-OneVision\footnote{\url{https://github.com/LLaVA-VL/LLaVA-NeXT}} and Cambrian\footnote{\url{https://github.com/cambrian-mllm/cambrian}} source codes.
Our experiments adopt two sizes of Qwen-Instruct LLMs:
\texttt{Qwen/Qwen2.5-7B-Instruct} and
\texttt{Qwen/Qwen3-4B-Instruct-2507}, paired with the SigLIP2 vision encoder
\texttt{google/siglip2-so400m-patch16-384}.
The multimodal projector is a 2-layer MLP with a GELU activation.
When applying Token Compression, the projector is augmented with a GELU followed by a $3{\times}3$ convolution layer with stride 2, which reduces the number of visual tokens by $4{\times}$.
For the training data, we use the Cambrian-Alignment dataset\footnote{\url{https://huggingface.co/datasets/nyu-visionx/Cambrian-Alignment}} for projector-only pretraining, and the Cambrian-7M\footnote{\url{https://huggingface.co/datasets/nyu-visionx/Cambrian-10M/blob/main/jsons/Cambrian7M_withsystemprompt.jsonl}} dataset and a randomly sampled subset of Cambrian-S-3M\footnote{\url{https://huggingface.co/datasets/nyu-visionx/Cambrian-S-3M}} for full-parameter fine-tuning. 
We follow the training hyperparameters from the standard scripts\footnote{\url{https://github.com/LLaVA-VL/LLaVA-NeXT/blob/main/scripts/train/pretrain_siglip.sh}}\textsuperscript{,}\footnote{\url{https://github.com/LLaVA-VL/LLaVA-NeXT/blob/main/scripts/train/finetune_si.sh}}. Specifically, the learning rate is set to $1{\times}10^{-3}$ for pretraining and $1{\times}10^{-5}$ for fine-tuning, with a batch size of 256. To implement the progressive $\mathrm{TopK}$ schedule, we define a region selection ratio that is gradually decreased over the course of training. The ratio starts at 100\% (i.e., all regions are selected) and is linearly reduced to 10\% by 60\% of the total training steps, after which it remains fixed.

\section{Evaluation Benchmarks}
\label{sec:evaluation-benchmarks}
\begin{table}[t]
\tablestyle{1pt}{1.1}
\resizebox{1.\linewidth}{!}{%
\begin{tabular}{x{80} x{70} x{100} x{120}}
\shline
\rowcolor[gray]{0.95}
Benchmark & Task & Domain & Citation \\
\hline
\rowcolor[gray]{0.95}
\multicolumn{4}{c}{\textit{Image Understanding Benchmarks}} \\
GQA & all & General & \mbox{\citep{hudson2019gqa}} \\
MME & perception & General & \mbox{\citep{fu2023mme}} \\
MMBench & all & General & \mbox{\citep{liu2023mmbench}} \\
MMVP & all & General & \mbox{\citep{eyeswideshut}} \\
RealWorldQA & all & General & \mbox{\citep{realworldqa}} \\
ADE & all & General & \mbox{\citep{cambrian}} \\
OmniBench & all & General & \mbox{\citep{cambrian}} \\
ScienceQA & image-based & Science \& Math & \mbox{\citep{lu2022learn}} \\
MMMU & vision & Science \& Math & \mbox{\citep{yue2023mmmu}} \\
MathVista & math & Science \& Math & \mbox{\citep{lu2023mathvista}} \\
ChartQA & all & Chart \& OCR & \mbox{\citep{masry2022chartqa}} \\
OCRBench & all & Chart \& OCR & \mbox{\citep{liu2023hidden}} \\
TextVQA & all & Chart \& OCR & \mbox{\citep{singh2019towards}} \\
\rowcolor[gray]{0.95}
\multicolumn{4}{c}{\textit{Video Understanding Benchmarks}} \\
VideoMME & all & Video & \mbox{\citep{fu2025videomme}} \\
MLVU & dev & Video & \mbox{\citep{zhou2024mlvu}} \\
EgoSchema & all & Video & \mbox{\citep{mangalam2023egoschema}} \\
LongVideoBench & val & Video & \mbox{\citep{wu2024longvideobench}} \\
NExT-QA & mc test & Video & \mbox{\citep{xiao2021nextqa}} \\
TempCompass & multi choice & Video & \mbox{\citep{liu2024tempcompass}} \\
\shline
\end{tabular}
}
\vspace{-.5em}
\caption{
\small
\textbf{Benchmarks used in the main evaluations.}
We evaluate MLLMs on 13 image benchmarks across three domains (General, Science \& Math, and Chart \& OCR) and 6 video benchmarks.
For each benchmark, we list the evaluated task or split, domain, and reference.
}
\vspace{-1.em}
\label{tab:benchmarks}
\end{table}

A list of evaluation benchmarks, along with their domain assignments and citations, is provided in \cref{tab:benchmarks}.
For image understanding tasks, we adopt the evaluation suite from Cambrian\footnote{\url{https://github.com/cambrian-mllm/cambrian/tree/main/eval}}, which consists of 13 benchmarks categorized into three domains: General, Science\,\&\,Math, and Chart\,\&\,OCR.
Since MME is scored on a 0–2000 scale unlike other benchmarks that range from 0 to 100, we divide the MME scores by 20 when computing the average to ensure a consistent scale across all benchmarks.
For video understanding tasks, we consider 6 benchmarks spanning diverse categories and video lengths, and employ the lmms-eval\footnote{\url{https://github.com/evolvinglmms-lab/lmms-eval}}~\citep{zhang2024lmmsevalrealitycheckevaluation} framework for evaluation.




\end{document}